\documentclass[11pt,a4paper]{article}
\usepackage[hyperref]{acl2020}
\usepackage{times}
\usepackage{latexsym}
\usepackage{xcolor}
\usepackage{multirow}
\usepackage{graphicx}
\usepackage{tabularx}
\usepackage{booktabs}
\usepackage{amsmath}
\usepackage{amssymb}
\usepackage{url}

\usepackage{flushend}
\usepackage{microtype}

\aclfinalcopy 
\def\aclpaperid{144} 

\title{{E}nhancing {W}ord {E}mbeddings with \\
{K}nowledge {E}xtracted from {L}exical {R}esources}

\author{Magdalena Biesialska\thanks{{ } Equal contribution} \qquad  Bardia Rafieian$^{*}$ \qquad  Marta R. Costa-juss\`a \\
TALP Research Center, Universitat Polit\`ecnica de Catalunya, Barcelona \\ \texttt{\{magdalena.biesialska,bardia.rafieian,marta.ruiz\}@upc.edu}}

\date{}
\hypersetup{draft}

\begin{document}
\maketitle
\begin{abstract}
In this work, we present an effective method for \textit{semantic specialization} of word vector representations. To this end, we use traditional word embeddings and apply specialization methods to better capture semantic relations between words. In our approach, we leverage external knowledge from rich lexical resources such as BabelNet. We also show that our proposed \textit{post-specialization} method based on an adversarial neural network with the Wasserstein distance allows to gain improvements over state-of-the-art methods on two tasks: word similarity and dialog state tracking.
\end{abstract}

\section{Introduction}
\label{introduction}
Vector representations of words (embeddings) have become the cornerstone of modern Natural Language Processing (NLP), as learning word vectors and utilizing them as features in downstream NLP tasks is the \textit{de facto} standard. Word embeddings \citep{Mikolov2013DistributedRO,pennington-etal-2014-glove} are typically trained in an unsupervised way on large monolingual corpora. Whilst such word representations are able to capture some syntactic as well as semantic information, their ability to map relations (e.g. synonymy, antonymy) between words is limited. To alleviate this deficiency, a set of refinement post-processing methods--called \textit{retrofitting} or \textit{semantic specialization}--has been introduced. In the next section, we discuss the intricacies of these methods in more detail.

To summarize, our contributions in this work are as follows:
\begin{itemize}
    \item We introduce a set of new linguistic constraints (i.e. synonyms and antonyms) created with BabelNet for three languages: English, German and Italian.
    \item We introduce an improved \textit{post-specialization} method (dubbed \textit{WGAN-postspec}), which demonstrates improved performance as compared to state-of-the-art \textit{DFFN} \citep{vulic-etal-2018-post} and \textit{AuxGAN} \citep{ponti-etal-2018-adversarial} models.
    \item We show that the proposed approach achieves performance improvements on an intrinsic task (word similarity) as well as on a downstream task (dialog state tracking).
\end{itemize}

\begin{figure*}
\includegraphics[width=1\textwidth]{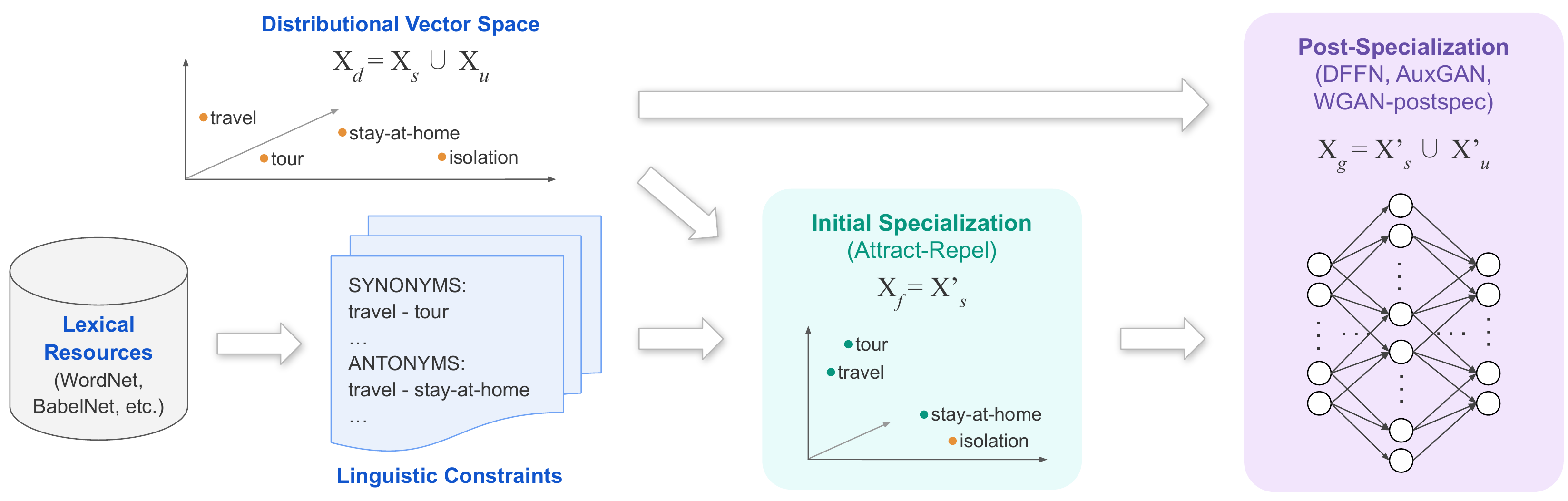}
\caption{\label{fig:approach}Illustration of the semantic specialization approach.}
\end{figure*}

\section{Related Work}
\label{related-work}
Numerous methods have been introduced for incorporating structured linguistic knowledge from external resources to word embeddings. Fundamentally, there exist three categories of \textit{semantic specialization} approaches: (a) \textit{joint methods} which incorporate lexical information during the training of distributional word vectors; (b) \textit{specialization} methods also referred to as \textit{retrofitting} methods which use post-processing techniques to inject semantic information from external lexical resources into pre-trained word vector representations; and (c) \textit{post-specialization} methods which use linguistic constraints to learn a general mapping function allowing to specialize the entire distributional vector space.

In general, \textit{joint methods} perform worse than the other two methods, and are not model-agnostic, as they are tightly coupled to the distributional word vector models (e.g. \textit{Word2Vec}, \textit{GloVe}). Therefore, in this work we concentrate on the \textit{specialization} and \textit{post-specialization} methods. Approaches which fall in the former category can be considered local specialization methods, where the most prominent examples are:
\textit{retrofitting} \citep{faruqui-etal-2015-retrofitting} which is a post-processing method to enrich word embeddings with knowledge from semantic lexicons, in this case it brings closer semantically similar words. \textit{Counter-fitting} \citep{mrksic-etal-2016-counter} likewise fine-tunes word representations; however, conversely to the \textit{retrofitting} technique it counter-fits the embeddings with respect to the given similarity and antonymy constraints.
\textit{Attract-Repel} \citep{mrksic-etal-2017-semantic} uses linguistic constraints obtained from external lexical resources to semantically specialize word embeddings. Similarly to \textit{counter-fitting} it injects synonymy and antonymy constraints into distributional word vector spaces. In contrast to \textit{counter-fitting}, this method does not ignore how updates of the example word vector pairs affect their relations to other word vectors.

On the other hand, the latter group, \textit{post-specialization} methods, performs global specialization of distributional spaces. We can distinguish: \textit{explicit retrofitting} \citep{glavas2018explicit} that was the first attempt to use external constraints (i.e. synonyms and antonyms) as training examples for learning an explicit mapping function for specializing the words not observed in the constraints. Later, a more robust \textit{DFFN} \citep{vulic-etal-2018-post} method was introduced with the same goal -- to specialize the full vocabulary by leveraging the already specialized subspace of seen words.

\section{Methodology}
\label{methodology}
In this paper, we propose an approach that builds upon previous works \citep{vulic-etal-2018-post,ponti-etal-2018-adversarial}. The process of specializing distributional vectors is a two-step procedure (as shown in Figure \ref{fig:approach}). First, an \textit{initial specialization} is performed (see §\ref{initial-spec}). In the second step, a global specialization mapping function is learned, allowing to generalize to unseen words (see §\ref{proposed-model}).

\subsection{Initial Specialization}
\label{initial-spec}
 In this step a subspace of distributional vectors for words that occur in the external constraints is specialized. To this end, fine-tuning of seen words can be performed using any \textit{specialization} method. In this work, we utilize \textit{Attract-Repel} model \citep{mrksic-etal-2017-semantic} as it offers state-of-the-art performance. This method allows to make use of both synonymy (\textit{attract}) and antonymy (\textit{repel}) constraints. More formally, given a set $\mathcal{A}$ of \textit{attract} word pairs and a set of $\mathcal{R}$ of \textit{repel} word pairs, let $\mathcal{V}_{\mathcal{S}}$ be the vocabulary of words seen in the constraints. Hence, each word pair $\left(v_{l}, v_{r}\right)$ is represented by a corresponding vector pair $\left(\mathbf{x}_{l}, \mathbf{x}_{r}\right)$. The model optimization method operates over mini-batches: a mini-batch $\mathcal{B}_{A}$ of synonymy pairs (of size $k_{1}$) and a mini-batch $\mathcal{B}_{R}$ of antonymy pairs (of size $k_{2}$). The pairs of negative examples $T_{A}\left(\mathcal{B}_{A}\right)=\left[\left(\mathbf{t}_{l}^{1}, \mathbf{t}_{r}^{1}\right), \ldots,\left(\mathbf{t}_{l}^{k_{1}}, \mathbf{t}_{r}^{k_{1}}\right)\right]$ and $T_{R}\left(\mathcal{B}_{R}\right)=\left[\left(\mathbf{t}_{l}^{1}, \mathbf{t}_{r}^{1}\right), \ldots,\left(\mathbf{t}_{l}^{k_{2}}, \mathbf{t}_{r}^{k_{2}}\right)\right]$ are drawn from $2\left(k_{1}+k_{2}\right)$ word vectors in $\mathcal{B}_{A} \cup \mathcal{B}_{R}$.

The negative examples serve the purpose of pulling synonym pairs closer and pushing antonym pairs further away with respect to their corresponding negative examples. For synonyms: 
\begin{align}\label{eq:ar-attract}
A\left(\mathcal{B}_{A}\right)= &\sum_{i=1}^{k_1}\left[\tau(\delta_{att}+\mathbf{x}_{l}^{i} \mathbf{t}_{l}^{i}-\mathbf{x}_{l}^{i} \mathbf{x}_{r}^{i}\right) + \nonumber \\ &+\tau\left(\delta_{att}+\mathbf{x}_{r}^{i} \mathbf{t}_{r}^{i}-\mathbf{x}_{l}^{i} \mathbf{x}_{r}^{i}\right)]
\end{align}
where $\tau$ is the rectifier function, and $\delta_{att}$ is the similarity margin determining the distance between synonymy vectors and how much closer they should be comparing to their negative examples. Similarly, the equation for antonyms is given as:
\begin{align}\label{eq:ar-repel}
R\left(\mathcal{B}_{R}\right)= &\sum_{i=1}^{k_2}\left[\tau(\delta_{rep}+\mathbf{x}_{l}^{i} \mathbf{x}_{r}^{i}-\mathbf{x}_{l}^{i} \mathbf{t}_{l}^{i}\right) + \nonumber \\ &+\tau\left(\delta_{rep}+\mathbf{x}_{l}^{i} \mathbf{x}_{r}^{i}-\mathbf{x}_{r}^{i} \mathbf{t}_{r}^{i}\right)]
\end{align}

A distributional regularization term is used to retain the quality of the original distributional vector space using $L_2$-regularization.
\begin{align}
Reg\left(\mathcal{B}_{A}, \mathcal{B}_{R}\right)=\sum_{\mathbf{x}_{i} \in V\left(\mathcal{B}_{A} \cup \mathcal{B}_{R}\right)} \lambda_{reg}\left\|\widehat{\mathbf{x}}_{i}-\mathbf{x}_{i}\right\|_{2}
\end{align}
where $\lambda_{reg}$ is a $L_2$-regularization constant, and $\widehat{x_i}$ is the original vector for the word ${x_i}$.

Consequently, the final cost function is formulated as follows:
\begin{align}\label{eq:ar-cost}
C(\mathcal{B}_{A},\mathcal{B}_{R})=A(\mathcal{B}_{A})+R(\mathcal{B}_{R})+Reg(\mathcal{B}_{A}, \mathcal{B}_{R})
\end{align}

\subsection{Proposed Post-Specialization Model}
\label{proposed-model}

Once the \textit{initial specialization} is completed, \textit{post-specialization} methods can be employed. This step is important, because local specialization affects only words seen in the constraints, and thus just a subset of the original distributional space $\mathbf{X}_d$. While \textit{post-specialization} methods learn a global specialization mapping function allowing them to generalize to unseen words $\mathbf{X}_u$. 

Given the specialized word vectors $\mathbf{X}^{\prime}_s$ from the vocabulary of seen words $\mathcal{V}_{\mathcal{S}}$, our proposed method propagates this signal to the entire distributional vector space using a generative adversarial network (GAN) \citep{goodfellow2014gan}. Hence, in our model, following the approach of \citet{ponti-etal-2018-adversarial}, we introduce adversarial losses. More specifically, the mapping function is learned through a combination of a standard $L_2$-loss with adversarial losses. The motivation behind this is to make the mappings more natural and ensure that vectors specialized for the full vocabulary are more realistic. To this end, we use the Wasserstein distance incorporated in the generative adversarial network (WGAN) \citep{arjovsky2017wgan} as well as its improved variant with gradient penalty (WGAN-GP) \citep{gulrajani2017wgan-gp}. For brevity, we call our model \textit{WGAN-postspec}, which is an umbrella term for the \textit{WGAN} and \textit{WGAN-GP} methods implemented in the proposed \textit{post-specialization} model. One of the benefits of using WGANs over vanilla GANs is that WGANs are generally more stable, and also they do not suffer from vanishing gradients.

Our proposed \textit{post-specialization} approach is based on the principles of GANs, as it is composed of two elements: a generator network $G$ and a discriminator network $D$. The gist of this concept, is to improve the generated samples through a min-max game between the generator and the discriminator.

In our \textit{post-specialization} model, a multi-layer feed-forward neural network, which trains a global mapping function, acts as the generator. Consequently, the generator is trained to produce predictions $G(\mathbf{x};\theta_{G})$ that are as similar as possible to the corresponding initially specialized word vectors ${\mathbf{x}^{\prime}_s}$. Therefore, a global mapping function is trained using word vector pairs, such that $\left(\mathbf{x}_{i}, \mathbf{x}_{i}^{\prime}\right)=\left\{\mathbf{x}_{i} \in \mathbf{X}_{s}, \mathbf{x}_{i}^{\prime} \in \mathbf{X}_{s}^{\prime}\right\}$. On the other hand, the discriminator $D(\mathbf{x};\theta_{D})$, which is a multi-layer classification network, tries to distinguish the generated samples from the initially specialized vectors sampled from ${\mathbf{X}^{\prime}_s}$. In this process, the differences between predictions and initially specialized vectors are used to improve the generator, resulting in more realistically looking outputs.

In general, for the GAN model we can define the loss $\mathcal{L}_{G}$ of the generator as:
\begin{align}\label{eq:lg-gan}
\mathcal{L}_{G}=&-\sum_{i=1}^{n} \log P(\text { spec }=1 | G(\mathbf{x}_{i} ; \theta_{G}) ; \theta_{D}) - \nonumber \\
&-\sum_{i=1}^{m} \log P(\text { spec }=0 | \mathbf{x}^{\prime}_{i} ; \theta_{D})
\end{align}
While the loss of the discriminator $\mathcal{L}_{D}$ is given as:
\begin{align}\label{eq:ld-gan}
\mathcal{L}_{D}=&-\sum_{i=1}^{n} \log P(\text { spec }=0 | G(\mathbf{x}_{i} ; \theta_{G}) ; \theta_{D}) - \nonumber \\
&-\sum_{i=1}^{m} \log P(\text { spec }=1 | \mathbf{x}^{\prime}_{i} ; \theta_{D})
\end{align}

In principle, the losses with Wasserstein distance can be formulated as follows:
\begin{align}\label{eq:lg-wgan}
\mathcal{L}_{G}=-\frac{1}{n}\sum_{i=1}^{n} D(G(\mathbf{x}_{i} ; \theta_{G}) ; \theta_{D})
\end{align}
and
\begin{align}\label{eq:ld-wgan}
\mathcal{L}_{D}=&\frac{1}{m}\sum_{i=1}^{m} D(\mathbf{x}^{\prime}_{i} ; \theta_{D}) - \nonumber \\ &-\frac{1}{n}\sum_{i=1}^{n} D(G(\mathbf{x}_{i} ; \theta_{G}) ; \theta_{D})
\end{align}

An alternative scenario with a gradient penalty (WGAN-GP) requires adding gradient penalty $\lambda$ coefficient in the Eq. (\ref{eq:ld-wgan}).

\section{Experiments}
\label{experiments}

\setlength{\tabcolsep}{0.7pt}
\begin{table*}
\centering
\def\arraystretch{1}
{\small
\begin{tabular}{lc rrr rrr rrr}
\toprule   
{} & {} & \multicolumn{3}{c}{\textbf{English}} & \multicolumn{3}{c}{\textbf{German}} & \multicolumn{3}{c}{\textbf{Italian}}\\
\cmidrule(lr){3-5}\cmidrule(lr){6-8}\cmidrule(lr){9-11}
{} & {} & \multirow{2}{*}{{\textit{overlap}}} & \multicolumn{1}{c}{\textit{disjoint}} & \multicolumn{1}{c}{\textit{disjoint}} & \multirow{2}{*}{{\textit{overlap}}} & 
\multicolumn{1}{c}{\textit{disjoint}} & \multicolumn{1}{c}{\textit{disjoint}} & \multirow{2}{*}{{\textit{overlap}}} & 
\multicolumn{1}{c}{\textit{disjoint}} & \multicolumn{1}{c}{\textit{disjoint}} \\
{} & {} & {} & \textit{simlex/verb} & \textit{wordsim} & & \textit{simlex/verb} & \textit{wordsim} & & \textit{simlex/verb} & \textit{wordsim}\\
\midrule
\multirow{2}{*}{\textbf{Synonyms}}
& \textit{babelnet} & 3,522,434 & 3,521,366 & 3,515,111 & 1,358,358 & 1,087,814 & 1,348,006 & 975,483 & 807,399 & 806,890 \\ 
& \textit{external + babelnet} & 4,545,045 & 4,396,350 & 3,515,111 & 1,360,040 & 1,089,338 & 1,349,612 & 976,877 & 808,605 & 808,225 \\
\cmidrule(r){2-11}
\multirow{2}{*}{\textbf{Antonyms}}
& \textit{babelnet} & 1,024 & 843 & 1,011 & 139 & 136 & 136 & 99 & 99 & 98 \\ 
& \textit{external + babelnet} & 381,777 & 352,099 & 378,365 & 1,823 & 1,662 & 1,744 & 883 & 769 & 851 \\
\bottomrule
\end{tabular}}
\caption{Number of synonym and antonym word pairs for English, German and Italian in two settings: \textit{babelnet}, \textit{external + babelnet}.}
\label{table:1}
\end{table*}

\paragraph{Pre-trained Word Embeddings.} In order to evaluate our proposed approach as well as to compare our results with respect to current state-of-the-art \textit{post-specialization} approaches, we use popular and readily available 300-dimensional pre-trained word vectors. \texttt{Word2Vec} \citep{Mikolov2013DistributedRO} embeddings for English were trained using skip-gram with negative sampling on the cleaned and tokenized Polyglot Wikipedia \citep{al-rfou-etal-2013-polyglot} by \citet{levy-goldberg-2014-dependency}, while German and Italian embeddings were trained using CBOW with negative sampling on WacKy corpora \citep{dinu2015improving,artetxe2017acl,artetxe2018aaai}. Moreover, \texttt{GloVe} vectors for English were trained on Common Crawl \citep{pennington-etal-2014-glove}.

\paragraph{Linguistic Constraints.} 
To perform \textit{semantic specialization} of word vector spaces, we exploit linguistic constraints used in previous works \citep{zhang-etal-2014-word,ono-etal-2015-word,vulic-etal-2018-post} (referred to as \textit{external}) as well as introduce a new set of constraints collected by us (referred to as \textit{babelnet}) for three languages: English, German and Italian. We use constraints in two different settings: \textit{disjoint} and \textit{overlap}. In the first setting, we remove all linguistic constraints that contain any of the words available in SimLex \citep{hill-etal-2015-simlex}, SimVerb \citep{gerz-etal-2016-simverb} and WordSim \citep{leviant2015separated} evaluation datasets. In the \textit{overlap} setting, we let the SimLex, SimVerb and WordSim words remain in the constraints. To summarize, we present the number of word pairs for English, German and Italian constraints in Table \ref{table:1}.

Let us discuss in more detail how the lists of constraints were constructed. In this work, we use two sets of linguistic constraints: \textit{external} and \textit{babelnet}. The first set of constraints was retrieved from WordNet \citep{Fellbaum1998WordNetAE} and Roget’s Thesaurus \citep{kipfer2009roget}, resulting in 1,023,082 synonymy and 380,873 antonymy word pairs. The second set of constraints, which is a part of our contribution, comprises synonyms and antonyms obtained using NASARI lexical embeddings \citep{camacho2016nasari} and BabelNet \citep{NavigliPonzetto:12aij}. As NASARI provides lexical information for BabelNet words in five languages (EN, ES, FR, DE and IT), we collected each word with its related BabelNetID (a sense database identifier) to extract the list of its synonyms and antonyms using BabelNet API.

Furthermore, to improve the list of Italian words, we also followed the approach proposed by \citet{SucameliL17}. The authors provided a new dataset of semantically related Italian word pairs. The dataset includes nouns, adjectives and verbs with their synonyms, antonyms and hypernyms. The information in this dataset was gathered by its authors through crowdsourcing from a pool of Italian native speakers. This way, we could concatenate Italian word pairs to provide a more complete list of synonyms and antonyms.
 
 Similarly, we refer to the work of \citet{scheible-schulte-im-walde-2014-database} that presents a new collection of semantically related word pairs in German, which was compiled through human evaluation. Relying on GermaNet and the respective JAVA API, the list of the word pairs was generated with a sampling technique. Finally, we used these word pairs in our experiments as external resources for the German language.

\paragraph{Initial Specialization and Post-Specialization.} Although, initially specialized vector spaces show gains over the non-specialized word embeddings, linguistic constraints represent only a fraction of their total vocabulary. Therefore, \textit{semantic specialization} is a two-step process. Firstly, we perform \textit{initial specialization} of the pre-trained word vectors by means of \textit{Attract-Repel} (see §\ref{related-work}) algorithm. The values of hyperparameter are set according to the default values: $\lambda_{reg}=10^{-9}, \delta_{sim}=0.6, \delta_{ant}=0.0$ and $k_1$ = $k_2$ = 50.
Afterward, to perform a specialization of the entire vocabulary, a global specialization mapping function is learned. In our \textit{WGAN-postspec} proposed approach, the \textit{post-specialization} model uses a GAN with improved loss functions by means of the Wasserstein distance and gradient penalty. Importantly, the optimization process differs depending on the algorithm implemented in our model. In the case of a vanilla GAN (\textit{AuxGAN}), standard stochastic gradient descent is used. While in the \textit{WGAN} model we employ RMSProp \citep{tieleman2012rmsprop}. Finally, in the case of the \textit{WGAN-GP}, Adam \citep{kingma2015adam} optimizer is applied.

\setlength{\tabcolsep}{5.8pt}
\begin{table*}[!htb]
\centering
\def\arraystretch{1}
{\footnotesize
\begin{tabular}{ll ccc ccc ccc ccc}
\toprule   
{} & {} & \multicolumn{12}{c}{\textbf{English}}\\
{} & {} & \multicolumn{6}{c}{\textsc{GloVe}} & \multicolumn{6}{c}{\textsc{Word2Vec}} \\
\cmidrule(lr){3-8}\cmidrule(lr){9-14}
{} & {} & \multicolumn{3}{c}{\textit{overlap}} & \multicolumn{3}{c}{\textit{disjoint}} & \multicolumn{3}{c}{\textit{overlap}} & \multicolumn{3}{c}{\textit{disjoint}} \\ 
\cmidrule(lr){3-5}\cmidrule(lr){6-8}\cmidrule(lr){9-11}\cmidrule(lr){12-14}
{} & {} & SL & SV & WS & SL & SV & WS & SL & SV & WS & SL & SV & WS \\
\midrule
\textsc{\textbf{original}} 
& {} & 0.407 & 0.280 & 0.655 & 0.407 & 0.280 & 0.655 & 0.414 & 0.272 & 0.593 & 0.414 & 0.272 & 0.593 \\
\cmidrule(lr){2-14}
\multirow{2}{*}{\textbf{\textsc{attract-}}} 
& \textit{a} & 0.781 & 0.761 & 0.597 & 0.407 & 0.280 & 0.655 & 0.778 & 0.761 & 0.574 & 0.414 & 0.272 & 0.593 \\ 
\multirow{2}{*}{\textbf{\textsc{repel}}} & \textit{b} & 0.407 & 0.282 & 0.655 & 0.407 & 0.282 & 0.655 & 0.414 & 0.275 & 0.594 & 0.414 & 0.275 & 0.593 \\ 
& \textit{c} & 0.784 & 0.763 & 0.595 & 0.407 & 0.282 & 0.655 & 0.776 & 0.763 & 0.560 & 0.414 & 0.275 & 0.593 \\
\cmidrule(lr){2-14}
\multirow{3}{*}{\textbf{\textsc{dffn}}} 
& \textit{a} & 0.785 & 0.764 & 0.600 & 0.645 & 0.531 & 0.678 & 0.781 & 0.763 & 0.571 & 0.553 & 0.430 & 0.593 \\ 
& \textit{b} & 0.699 & 0.562 & 0.703 & 0.458 & 0.324 & 0.679 & 0.351 & 0.237 & 0.506 & 0.387 & 0.245 & 0.578 \\ 
& \textit{c} & 0.783 & 0.764 & 0.597 & 0.646 & 0.535 & 0.684 & 0.777 & 0.763 & 0.560 & 0.538 & 0.381 & 0.594 \\
\cmidrule(lr){2-14}
\multirow{3}{*}{\textbf{\textsc{auxgan}}} 
& \textit{a} & 0.789 & 0.764 & 0.659 & 0.652 & 0.552 & 0.642 & 0.782 & 0.762 & 0.550 & 0.581 & 0.434 & 0.602 \\ 
& \textit{b} & 0.734 & 0.647 & 0.627 & 0.417 & 0.284 & 0.658 & 0.405 & 0.269 & 0.587 & 0.395 & 0.260 & 0.581 \\ 
& \textit{c} & 0.796 & 0.767 & 0.639 & 0.659 & 0.560 & 0.669 & 0.782 & 0.755 & 0.588 & 0.583 & 0.438 & 0.603 \\
\cmidrule(lr){2-14}
\multirow{3}{*}{\textbf{\textsc{wgan}}} 
& \textit{a} & 0.809 & 0.767 & 0.652 & 0.661 & 0.553 & 0.642 & 0.780 & 0.749 & 0.602 & 0.580 & 0.446 & 0.608 \\ 
& \textit{b} & 0.722 & 0.635 & 0.654 & 0.452 & 0.279 & 0.671 & 0.392 & 0.262 & 0.590 & 0.397 & 0.269 & 0.580 \\ 
& \textit{c} & 0.808 & 0.765 & 0.653 & 0.663 & 0.549 & 0.665 & 0.771 & 0.737 & 0.614 & 0.586 & 0.440 & 0.611 \\
\cmidrule(lr){2-14}
\multirow{3}{*}{\textbf{\textsc{wgan-gp}}} 
& \textit{a} & 0.810 & 0.751 & 0.669 & 0.660 & 0.548 & 0.669 & 0.776 & 0.742 & 0.600 & 0.586 & 0.462 & 0.605 \\ 
& \textit{b} & 0.722 & 0.622 & 0.646 & 0.461 & 0.282 & 0.676 & 0.396 & 0.254 & 0.567 & 0.398 & 0.267 & 0.581 \\ 
& \textit{c} & 0.798 & 0.732 & 0.715 & 0.660 & 0.551 & 0.672 & 0.775 & 0.614 & 0.590 & 0.585 & 0.463 & 0.609 \\
\bottomrule
\end{tabular}}
\caption{Spearman's $\rho$ correlation scores on SimLex-999 (SL), SimVerb-3500 (SV) and WordSim-353 (WS). Evaluation was performed using constraints in three settings: \textit{(a) external}, \textit{(b) babelnet}, \textit{(c) external + babelnet}.}
\label{tab:wordsim-en-results}
\end{table*}

\setlength{\tabcolsep}{0.6pt}
\begin{table}[!htb]
\centering
\def\arraystretch{1}
{\footnotesize
\begin{tabular}{ll cccc cccc}
\toprule   
{} & {} & \multicolumn{4}{c}{\textbf{German}} & \multicolumn{4}{c}{\textbf{Italian}} \\
{} & {} & \multicolumn{4}{c}{\textsc{Word2Vec}} & \multicolumn{4}{c}{\textsc{Word2Vec}} \\
\cmidrule(lr){3-6}\cmidrule(lr){7-10}
{} & {} & \multicolumn{2}{c}{\textit{overlap}} & \multicolumn{2}{c}{\textit{disjoint}} & \multicolumn{2}{c}{\textit{overlap}} & \multicolumn{2}{c}{\textit{disjoint}} \\ 
\cmidrule(lr){3-4}\cmidrule(lr){5-6}\cmidrule(lr){7-8}\cmidrule(lr){9-10}
{} & {} & SL & WS & SL & WS & SL & WS & SL & WS \\
\midrule
\textsc{\textbf{original}} & {} & 0.358 & 0.538 & 0.358 & 0.538 & 0.356 & 0.563 & 0.356 & 0.563 \\
\cmidrule(lr){2-10}
\multirow{2}{*}{\textbf{\textsc{attract-}}} & \textit{a} & 0.360 & 0.537 & 0.358 & 0.538 & 0.376 & 0.568 & 0.364 & 0.565 \\
\multirow{2}{*}{\textbf{\textsc{repel}}} & \textit{b} & 0.358 & 0.538 & 
0.358 & 0.538 & 0.366 & 0.568 & 0.366 & 0.559 \\ 
& \textit{c} & 0.360 & 0.538 & 0.358 & 0.538 & 0.378 & 0.566 & 0.367 & 0.564 \\
\cmidrule(lr){2-10}
\multirow{3}{*}{\textbf{\textsc{dffn}}} & \textit{a} & 0.366 & 0.422 & 0.370 & 0.452 & 0.381 & 0.512 & 0.365 & 0.519 \\ 
& \textit{b} & 0.354 & 0.538 & 0.348 & 0.538 & 0.364 & 0.559 & 0.361 & 0.560 \\ 
& \textit{c} & 0.359 & 0.541 & 0.358 & 0.533 & 0.376 & 0.561 & 0.369 & 0.559 \\
\cmidrule(lr){2-10}
\multirow{3}{*}{\textbf{\textsc{auxgan}}} 
& \textit{a} & 0.331 & 0.532 & 0.325 & 0.535 & 0.362 & 0.561 & 0.348 & 0.560 \\ 
& \textit{b} & 0.369 & 0.552 & 0.373 & 0.561 & 0.361 & 0.559 & 0.364 & 0.563 \\ 
& \textit{c} & 0.369 & 0.564 & 0.365 & 0.556 & 0.365 & 0.566 & 0.368 & 0.563 \\
\cmidrule(lr){2-10}
\multirow{3}{*}{\textbf{\textsc{wgan}}} 
& \textit{a} & 0.331 & 0.528 & 0.327 & 0.531 & 0.361 & 0.558 & 0.344 & 0.558 \\ 
& \textit{b} & 0.364 & 0.558 & 0.367 & 0.559 & 0.359 & 0.553 & 0.367 & 0.559 \\ 
& \textit{c} & 0.371 & 0.559 & 0.364 & 0.560 & 0.367 & 0.567 & 0.370 & 0.562 \\
\bottomrule
\end{tabular}}
\caption{Spearman's $\rho$ correlation scores on SimLex-999 (SL) and WordSim-353 (WS). Evaluation was performed using constraints in three settings: \textit{(a) external}, \textit{(b) babelnet}, \textit{(c) external + babelnet}.}
\label{tab:wordsim-de-it-results}
\end{table}

\begin{table}[!htb]
\centering
\def\arraystretch{1}
{\footnotesize
\begin{tabular}{lc}
\toprule
 & \textsc{GloVe} \\
\midrule
\textsc{Original} & 0.797 \\
\textsc{Attract-Repel} & 0.817 \\
\textsc{DFFN} & 0.829 \\
\textsc{AuxGAN} & 0.836 \\
\textsc{WGAN-postspec} & 0.838 \\
\bottomrule
\end{tabular}}
\caption{DST results for English.}
\label{tab:dst-results}
\end{table}

\section{Results}
\label{results}
\subsection{Word Similarity}
\label{word-similarity}

We report our experimental results with respect to a common intrinsic word similarity task, using standard benchmarks: SimLex-999 and WordSim-353 for English, German and Italian, as well as SimVerb-3500 for English. Each dataset contains human similarity ratings, and we evaluate the similarity measure using the Spearman's $\rho$ rank correlation coefficient. In Table \ref{tab:wordsim-en-results}, we present results for English benchmarks, whereas results for German and Italian are reported in Table \ref{tab:wordsim-de-it-results}.

Word embeddings are evaluated in two scenarios: \textit{disjoint} where words observed in the benchmark datasets are removed from the linguistic constraints; and \textit{overlap} where all words provided in the linguistic constraints are utilized. We use the \textit{overlap} setting in a downstream task (see §\ref{dst}). In the tasks we report scores for \textit{Original} (non-specialized) word vectors, \textit{initial specialization} method \textit{Attract-Repel} \citep{mrksic-etal-2017-semantic}, and three post-specialization methods: \textit{DFFN} \citep{vulic-etal-2018-post}, \textit{AuxGAN} \citep{ponti-etal-2018-adversarial} and our proposed model \textit{WGAN-postspec} (in two scenarios: \textit{WGAN} and \textit{WGAN-GP}).

The results suggest that the \textit{post-specialization} methods bring improvements in the specialization of the distributional word vector space. Overall, the highest correlation scores are reported for the models with adversarial losses. We also observe that the proposed \textit{WGAN-postspec} achieves fairly consistent correlation gains with \textsc{GloVe} vectors on the SimLex dataset. Interestingly, while exploiting additional constraints (i.e. \textit{external + babelnet}) generally boosts correlation scores for German and Italian, the results are not conclusive in the case of English, and thus they require further investigation.

\subsection{Dialog State Tracking}
\label{dst}
We also evaluate our proposed approach on a dialog state tracking (DST) downstream task. This task is a standard language understanding task, which allows to differentiate between word similarity and relatedness. To perform the evaluation we follow previous works \citep{henderson-etal-2014-second,Williams2016TheDS,mrksic-etal-2017-semantic}. Concretely, a DST model computes probability based only on pre-trained word embeddings. We use Wizard-of-Oz (WOZ) v.2.0 dataset \citep{wen16,mrksic-etal-2017-neural} composed of 600 training dialogues as well as 200 development and 400 test dialogues. 

In our experiments, we report results with a standard \textit{joint goal accuracy} (JGA) score. The results in Table \ref{tab:dst-results} confirm our findings from the previous word similarity task, as initial semantic \textit{specialization} and \textit{post-specialization} (in particular \textit{WGAN-postspec}) yield improvements over original distributional word vectors. We expect this conclusion to hold in all settings; however, additional experiments for different languages and word embeddings would be beneficial.

\section{Conclusion and Future Work}
\label{conclusion-and-future-work}
In this work, we presented a method to perform \textit{semantic specialization} of word vectors. Specifically, we compiled a new set of constraints obtained from BabelNet. Moreover, we improved a state-of-the-art \textit{post-specialization} method by incorporating adversarial losses with the Wasserstein distance. Our results obtained in an intrinsic and an extrinsic task, suggest that our method yields performance gains over current methods.

In the future, we plan to introduce constraints for asymmetric relations as well as extend our proposed method to leverage them. Moreover, we plan to experiment with adapting our model to a multilingual scenario, to be able to use it in a neural machine translation task. We make the code and resources available at: \url{https://github.com/mbiesialska/wgan-postspec}

\section*{Acknowledgments}
 
We thank the anonymous reviewers for their insightful comments.
This work is supported in part by the Spanish Ministerio de Econom\'ia y Competitividad, the European Regional Development Fund through the  postdoctoral  senior grant Ram\'on y Cajal and by the Agencia  Estatal  de  Investigaci\'on through the projects EUR2019-103819 and PCIN-2017-079.

\bibliography{acl2020}

\begin{thebibliography}{33}
\expandafter\ifx\csname natexlab\endcsname\relax\def\natexlab#1{#1}\fi

\bibitem[{Al-Rfou{'} et~al.(2013)Al-Rfou{'}, Perozzi, and
  Skiena}]{al-rfou-etal-2013-polyglot}
Rami Al-Rfou{'}, Bryan Perozzi, and Steven Skiena. 2013.
\newblock \href {https://www.aclweb.org/anthology/W13-3520} {{P}olyglot:
  Distributed word representations for multilingual {NLP}}.
\newblock In \emph{Proceedings of the Seventeenth Conference on Computational
  Natural Language Learning}, pages 183--192, Sofia, Bulgaria. Association for
  Computational Linguistics.

\bibitem[{Arjovsky et~al.(2017)Arjovsky, Chintala, and
  Bottou}]{arjovsky2017wgan}
Martin Arjovsky, Soumith Chintala, and L\'{e}on Bottou. 2017.
\newblock Wasserstein generative adversarial networks.
\newblock In \emph{Proceedings of the 34th International Conference on Machine
  Learning - Volume 70}, ICML’17, page 214–223. JMLR.org.

\bibitem[{Artetxe et~al.(2017)Artetxe, Labaka, and Agirre}]{artetxe2017acl}
Mikel Artetxe, Gorka Labaka, and Eneko Agirre. 2017.
\newblock Learning bilingual word embeddings with (almost) no bilingual data.
\newblock In \emph{Proceedings of the 55th Annual Meeting of the Association
  for Computational Linguistics (Volume 1: Long Papers)}, pages 451--462.

\bibitem[{Artetxe et~al.(2018)Artetxe, Labaka, and Agirre}]{artetxe2018aaai}
Mikel Artetxe, Gorka Labaka, and Eneko Agirre. 2018.
\newblock Generalizing and improving bilingual word embedding mappings with a
  multi-step framework of linear transformations.
\newblock In \emph{Proceedings of the Thirty-Second AAAI Conference on
  Artificial Intelligence}, pages 5012--5019.

\bibitem[{Camacho-Collados et~al.(2016)Camacho-Collados, Pilehvar, and
  Navigli}]{camacho2016nasari}
Jos{\'e} Camacho-Collados, Mohammad~Taher Pilehvar, and Roberto Navigli. 2016.
\newblock Nasari: Integrating explicit knowledge and corpus statistics for a
  multilingual representation of concepts and entities.
\newblock \emph{Artificial Intelligence}, 240:36--64.

\bibitem[{Dinu et~al.(2015)Dinu, Lazaridou, and Baroni}]{dinu2015improving}
Georgiana Dinu, Angeliki Lazaridou, and Marco Baroni. 2015.
\newblock Improving zero-shot learning by mitigating the hubness problem.
\newblock \emph{Proceedings of ICLR}.

\bibitem[{Faruqui et~al.(2015)Faruqui, Dodge, Jauhar, Dyer, Hovy, and
  Smith}]{faruqui-etal-2015-retrofitting}
Manaal Faruqui, Jesse Dodge, Sujay~Kumar Jauhar, Chris Dyer, Eduard Hovy, and
  Noah~A. Smith. 2015.
\newblock \href {https://doi.org/10.3115/v1/N15-1184} {Retrofitting word
  vectors to semantic lexicons}.
\newblock In \emph{Proceedings of the 2015 Conference of the North {A}merican
  Chapter of the Association for Computational Linguistics: Human Language
  Technologies}, pages 1606--1615, Denver, Colorado. Association for
  Computational Linguistics.

\bibitem[{Fellbaum(1998)}]{Fellbaum1998WordNetAE}
Christiane Fellbaum. 1998.
\newblock Wordnet: An electronic lexical database mit press.

\bibitem[{Gerz et~al.(2016)Gerz, Vuli{\'c}, Hill, Reichart, and
  Korhonen}]{gerz-etal-2016-simverb}
Daniela Gerz, Ivan Vuli{\'c}, Felix Hill, Roi Reichart, and Anna Korhonen.
  2016.
\newblock \href {https://doi.org/10.18653/v1/D16-1235} {{S}im{V}erb-3500: A
  large-scale evaluation set of verb similarity}.
\newblock In \emph{Proceedings of the 2016 Conference on Empirical Methods in
  Natural Language Processing}, pages 2173--2182, Austin, Texas. Association
  for Computational Linguistics.

\bibitem[{Glava{\v{s}} and Vuli{\'c }(2018)}]{glavas2018explicit}
Goran Glava{\v{s}} and {I.} Vuli{\'c }. 2018.
\newblock \href {https://doi.org/10.18653/v1/P18-1004} {Explicit retrofitting
  of distributional word vectors}.
\newblock In \emph{Proceedings of the 56th Annual Meeting of the Association
  for Computational Linguistics (Volume 1: Long Papers)}, pages 34--45,
  Melbourne, Australia. Association for Computational Linguistics.

\bibitem[{Goodfellow et~al.(2014)Goodfellow, Pouget-Abadie, Mirza, Xu,
  Warde-Farley, Ozair, Courville, and Bengio}]{goodfellow2014gan}
Ian~J. Goodfellow, Jean Pouget-Abadie, Mehdi Mirza, Bing Xu, David
  Warde-Farley, Sherjil Ozair, Aaron Courville, and Yoshua Bengio. 2014.
\newblock Generative adversarial nets.
\newblock In \emph{Proceedings of the 27th International Conference on Neural
  Information Processing Systems - Volume 2}, NIPS’14, page 2672–2680,
  Cambridge, MA, USA. MIT Press.

\bibitem[{Gulrajani et~al.(2017)Gulrajani, Ahmed, Arjovsky, Dumoulin, and
  Courville}]{gulrajani2017wgan-gp}
Ishaan Gulrajani, Faruk Ahmed, Martin Arjovsky, Vincent Dumoulin, and Aaron~C
  Courville. 2017.
\newblock \href
  {http://papers.nips.cc/paper/7159-improved-training-of-wasserstein-gans.pdf}
  {Improved training of wasserstein gans}.
\newblock In I.~Guyon, U.~V. Luxburg, S.~Bengio, H.~Wallach, R.~Fergus,
  S.~Vishwanathan, and R.~Garnett, editors, \emph{Advances in Neural
  Information Processing Systems 30}, pages 5767--5777. Curran Associates, Inc.

\bibitem[{Henderson et~al.(2014)Henderson, Thomson, and
  Williams}]{henderson-etal-2014-second}
Matthew Henderson, Blaise Thomson, and Jason~D. Williams. 2014.
\newblock \href {https://doi.org/10.3115/v1/W14-4337} {The second dialog state
  tracking challenge}.
\newblock In \emph{Proceedings of the 15th Annual Meeting of the Special
  Interest Group on Discourse and Dialogue ({SIGDIAL})}, pages 263--272,
  Philadelphia, PA, U.S.A. Association for Computational Linguistics.

\bibitem[{Hill et~al.(2015)Hill, Reichart, and
  Korhonen}]{hill-etal-2015-simlex}
Felix Hill, Roi Reichart, and Anna Korhonen. 2015.
\newblock \href {https://doi.org/10.1162/COLI_a_00237} {{S}im{L}ex-999:
  Evaluating semantic models with (genuine) similarity estimation}.
\newblock \emph{Computational Linguistics}, 41(4):665--695.

\bibitem[{Kingma and Ba(2015)}]{kingma2015adam}
Diederik~P. Kingma and Jimmy Ba. 2015.
\newblock Adam: {A} method for stochastic optimization.
\newblock In \emph{3rd International Conference on Learning Representations,
  {ICLR} 2015, San Diego, CA, USA, May 7-9, 2015, Conference Track
  Proceedings}.

\bibitem[{Kipfer(2009)}]{kipfer2009roget}
B.A. Kipfer. 2009.
\newblock \emph{Roget's 21st Century Thesaurus (3rd Edition)}.
\newblock Philip Lief Group.

\bibitem[{Leviant and Reichart(2015)}]{leviant2015separated}
Ira Leviant and Roi Reichart. 2015.
\newblock \href {http://arxiv.org/abs/1508.00106} {Separated by an un-common
  language: Towards judgment language informed vector space modeling}.

\bibitem[{Levy and Goldberg(2014)}]{levy-goldberg-2014-dependency}
Omer Levy and Yoav Goldberg. 2014.
\newblock \href {https://doi.org/10.3115/v1/P14-2050} {Dependency-based word
  embeddings}.
\newblock In \emph{Proceedings of the 52nd Annual Meeting of the Association
  for Computational Linguistics (Volume 2: Short Papers)}, pages 302--308,
  Baltimore, Maryland. Association for Computational Linguistics.

\bibitem[{Mikolov et~al.(2013)Mikolov, Sutskever, Chen, Corrado, and
  Dean}]{Mikolov2013DistributedRO}
Tomas Mikolov, Ilya Sutskever, Kai Chen, Gregory~S. Corrado, and Jeffrey Dean.
  2013.
\newblock Distributed representations of words and phrases and their
  compositionality.
\newblock In \emph{NIPS}.

\bibitem[{Mrk{\v{s}}i{\'c} et~al.(2016)Mrk{\v{s}}i{\'c}, {\'O}~S{\'e}aghdha,
  Thomson, Ga{\v{s}}i{\'c}, Rojas-Barahona, Su, Vandyke, Wen, and
  Young}]{mrksic-etal-2016-counter}
Nikola Mrk{\v{s}}i{\'c}, Diarmuid {\'O}~S{\'e}aghdha, Blaise Thomson, Milica
  Ga{\v{s}}i{\'c}, Lina~M. Rojas-Barahona, Pei-Hao Su, David Vandyke,
  Tsung-Hsien Wen, and Steve Young. 2016.
\newblock \href {https://doi.org/10.18653/v1/N16-1018} {Counter-fitting word
  vectors to linguistic constraints}.
\newblock In \emph{Proceedings of the 2016 Conference of the North {A}merican
  Chapter of the Association for Computational Linguistics: Human Language
  Technologies}, pages 142--148, San Diego, California. Association for
  Computational Linguistics.

\bibitem[{Mrk{\v{s}}i{\'c} et~al.(2017{\natexlab{a}})Mrk{\v{s}}i{\'c},
  {\'O}~S{\'e}aghdha, Wen, Thomson, and Young}]{mrksic-etal-2017-neural}
Nikola Mrk{\v{s}}i{\'c}, Diarmuid {\'O}~S{\'e}aghdha, Tsung-Hsien Wen, Blaise
  Thomson, and Steve Young. 2017{\natexlab{a}}.
\newblock \href {https://doi.org/10.18653/v1/P17-1163} {Neural belief tracker:
  Data-driven dialogue state tracking}.
\newblock In \emph{Proceedings of the 55th Annual Meeting of the Association
  for Computational Linguistics (Volume 1: Long Papers)}, pages 1777--1788,
  Vancouver, Canada. Association for Computational Linguistics.

\bibitem[{Mrk{\v{s}}i{\'c} et~al.(2017{\natexlab{b}})Mrk{\v{s}}i{\'c},
  Vuli{\'c}, {\'O}~S{\'e}aghdha, Leviant, Reichart, Ga{\v{s}}i{\'c}, Korhonen,
  and Young}]{mrksic-etal-2017-semantic}
Nikola Mrk{\v{s}}i{\'c}, Ivan Vuli{\'c}, Diarmuid {\'O}~S{\'e}aghdha, Ira
  Leviant, Roi Reichart, Milica Ga{\v{s}}i{\'c}, Anna Korhonen, and Steve
  Young. 2017{\natexlab{b}}.
\newblock \href {https://doi.org/10.1162/tacl_a_00063} {Semantic specialization
  of distributional word vector spaces using monolingual and cross-lingual
  constraints}.
\newblock \emph{Transactions of the Association for Computational Linguistics},
  5:309--324.

\bibitem[{Navigli and Ponzetto(2012)}]{NavigliPonzetto:12aij}
Roberto Navigli and Simone~Paolo Ponzetto. 2012.
\newblock {B}abel{N}et: {T}he automatic construction, evaluation and
  application of a wide-coverage multilingual semantic network.
\newblock \emph{Artificial Intelligence}, 193:217--250.

\bibitem[{Ono et~al.(2015)Ono, Miwa, and Sasaki}]{ono-etal-2015-word}
Masataka Ono, Makoto Miwa, and Yutaka Sasaki. 2015.
\newblock \href {https://doi.org/10.3115/v1/N15-1100} {Word embedding-based
  antonym detection using thesauri and distributional information}.
\newblock In \emph{Proceedings of the 2015 Conference of the North {A}merican
  Chapter of the Association for Computational Linguistics: Human Language
  Technologies}, pages 984--989, Denver, Colorado. Association for
  Computational Linguistics.

\bibitem[{Pennington et~al.(2014)Pennington, Socher, and
  Manning}]{pennington-etal-2014-glove}
Jeffrey Pennington, Richard Socher, and Christopher Manning. 2014.
\newblock \href {https://doi.org/10.3115/v1/D14-1162} {{G}love: Global vectors
  for word representation}.
\newblock In \emph{Proceedings of the 2014 Conference on Empirical Methods in
  Natural Language Processing ({EMNLP})}, pages 1532--1543, Doha, Qatar.
  Association for Computational Linguistics.

\bibitem[{Ponti et~al.(2018)Ponti, Vuli{\'c}, Glava{\v{s}}, Mrk{\v{s}}i{\'c},
  and Korhonen}]{ponti-etal-2018-adversarial}
Edoardo~Maria Ponti, Ivan Vuli{\'c}, Goran Glava{\v{s}}, Nikola
  Mrk{\v{s}}i{\'c}, and Anna Korhonen. 2018.
\newblock \href {https://doi.org/10.18653/v1/D18-1026} {Adversarial propagation
  and zero-shot cross-lingual transfer of word vector specialization}.
\newblock In \emph{Proceedings of the 2018 Conference on Empirical Methods in
  Natural Language Processing}, pages 282--293, Brussels, Belgium. Association
  for Computational Linguistics.

\bibitem[{Scheible and Schulte~im
  Walde(2014)}]{scheible-schulte-im-walde-2014-database}
Silke Scheible and Sabine Schulte~im Walde. 2014.
\newblock \href {https://doi.org/10.3115/v1/W14-5814} {A database of
  paradigmatic semantic relation pairs for {G}erman nouns, verbs, and
  adjectives}.
\newblock In \emph{Proceedings of Workshop on Lexical and Grammatical Resources
  for Language Processing}, pages 111--119, Dublin, Ireland. Association for
  Computational Linguistics and Dublin City University.

\bibitem[{Sucameli and Lenci(2017)}]{SucameliL17}
Irene Sucameli and Alessandro Lenci. 2017.
\newblock \href {http://ceur-ws.org/Vol-2006/paper041.pdf} {Parad-it: Eliciting
  italian paradigmatic relations with crowdsourcing}.
\newblock In \emph{Proceedings of the Fourth Italian Conference on
  Computational Linguistics (CLiC-it 2017), Rome, Italy, December 11-13, 2017}.

\bibitem[{Tieleman and Hinton(2012)}]{tieleman2012rmsprop}
Tijmen Tieleman and Geoffrey Hinton. 2012.
\newblock Lecture 6.5-rmsprop: Divide the gradient by a running average of its
  recent magnitude.
\newblock \emph{COURSERA: Neural networks for machine learning}, 4(2):26--31.

\bibitem[{Vuli{\'c} et~al.(2018)Vuli{\'c}, Glava{\v{s}}, Mrk{\v{s}}i{\'c}, and
  Korhonen}]{vulic-etal-2018-post}
Ivan Vuli{\'c}, Goran Glava{\v{s}}, Nikola Mrk{\v{s}}i{\'c}, and Anna Korhonen.
  2018.
\newblock \href {https://doi.org/10.18653/v1/N18-1048} {Post-specialisation:
  Retrofitting vectors of words unseen in lexical resources}.
\newblock In \emph{Proceedings of the 2018 Conference of the North {A}merican
  Chapter of the Association for Computational Linguistics: Human Language
  Technologies, Volume 1 (Long Papers)}, pages 516--527, New Orleans,
  Louisiana. Association for Computational Linguistics.

\bibitem[{Wen et~al.(2017)Wen, Vandyke, Mrk{\v{s}}i{\'c}, Ga{\v{s}}i{\'c},
  Rojas-Barahona, Su, Ultes, and Young}]{wen16}
Tsung-Hsien Wen, David Vandyke, Nikola Mrk{\v{s}}i{\'c}, Milica
  Ga{\v{s}}i{\'c}, Lina~M. Rojas-Barahona, Pei-Hao Su, Stefan Ultes, and Steve
  Young. 2017.
\newblock \href {https://www.aclweb.org/anthology/E17-1042} {A network-based
  end-to-end trainable task-oriented dialogue system}.
\newblock In \emph{Proceedings of the 15th Conference of the {E}uropean Chapter
  of the Association for Computational Linguistics: Volume 1, Long Papers},
  pages 438--449, Valencia, Spain. Association for Computational Linguistics.

\bibitem[{Williams et~al.(2016)Williams, Raux, and
  Henderson}]{Williams2016TheDS}
Jason~D. Williams, Antoine Raux, and Matthew Henderson. 2016.
\newblock The dialog state tracking challenge series: A review.
\newblock \emph{Dialogue \& Discourse}, 7:4--33.

\bibitem[{Zhang et~al.(2014)Zhang, Salwen, Glass, and
  Gliozzo}]{zhang-etal-2014-word}
Jingwei Zhang, Jeremy Salwen, Michael Glass, and Alfio Gliozzo. 2014.
\newblock \href {https://doi.org/10.3115/v1/D14-1161} {Word semantic
  representations using {B}ayesian probabilistic tensor factorization}.
\newblock In \emph{Proceedings of the 2014 Conference on Empirical Methods in
  Natural Language Processing ({EMNLP})}, pages 1522--1531, Doha, Qatar.
  Association for Computational Linguistics.

\end{thebibliography}
\bibliographystyle{acl_natbib}

\end{document}